\pdfoutput=1
\documentclass[11pt]{article}

\usepackage[final]{acl}

\usepackage{times}
\usepackage{latexsym}

\usepackage[T1]{fontenc}
\usepackage[utf8]{inputenc}

\usepackage{microtype}

\usepackage{inconsolata}

\usepackage{graphicx}

\usepackage{booktabs, multirow} % for borders and merged ranges
\usepackage{soul}% for underlines
\usepackage{changepage,threeparttable} % for wide tables
\usepackage{fancyvrb}
\usepackage{tcolorbox}
\usepackage{cleveref}
\usepackage{float}
\usepackage[russian,english]{babel}
\crefformat{section}{\S#2#1#3}

\title{CUTE: Measuring LLMs' Understanding of Their Tokens}
\author{Lukas Edman$^{1,3}$ \qquad Helmut Schmid$^{1}$ \qquad Alexander Fraser$^{2,3,4}$ \vspace{.2cm}\\ 
$^{1}$Center for Information and Language Processing, LMU Munich \\ 
$^{2}$School of Computation, Information and Technology, TU Munich \\
$^{3}$Munich Center for Machine Learning \\ 
$^{4}$Munich Data Science Institute \\ 
\vspace{.1cm} {\tt \small lukas@cis.lmu.de, schmid@cis.lmu.de}
}

\begin{document}
\maketitle

\begin{abstract}
Large Language Models (LLMs) show remarkable performance on a wide variety of tasks. Most LLMs split text into multi-character tokens and process them as atomic units without direct access to individual characters. This raises the question: To what extent can LLMs learn orthographic information?  To answer this, we propose a new benchmark, CUTE, which features a collection of tasks designed to test the orthographic knowledge of LLMs. We evaluate popular LLMs on CUTE, finding that most of them seem to know the spelling of their tokens, yet fail to use this information effectively to manipulate text, calling into question how much of this knowledge is generalizable. 
\end{abstract}

\section{Introduction}

Large Language Models (LLMs) attract a lot of interest %in recent years 
due to their strong performance on many NLP tasks. They have demonstrated a level of fluency rivaling humans.
% with English that often makes it difficult to tell them apart from humans.
However, 
it is often overlooked that LLMs
%there is an element of LLMs that is often overlooked: they have no 
lack direct access to the characters composing their tokens. 
They can only infer knowledge about the characters from the context during pretraining or instruction tuning.
%They must learn this from context via their pretraining, or indirectly in the form of instruction tuning. 
While there are models that use characters as input units, none of them have been instruction-tuned (to our knowledge).
%And while there are models using characters as tokens, none have been developed into instruction-tuned LLMs (to our knowledge).

Our work examines how well LLMs understand %In this work, we aim to gauge LLMs' ability to comprehend 
the composition of their tokens. 
This knowledge enables LLMs to better generalize to new languages 
and to perform well on a variety of tasks involving character-level understanding.
Tasks such as word puzzles, poetry generation (e.g. alliterations), or parsing ciphers all require very explicit use of characters to achieve. More popular tasks such as code completion, morphological inflection, or spelling correction also require character-level information to a lesser extent, though these tasks also require semantic knowledge, which we wish to ablate.

% (e.g. code, crosswords, poetry, etc.), or any task requiring arbitrary
% string operations. %operation on arbitrary strings.

We introduce \textbf{C}haracter-level \textbf{U}nderstanding of \textbf{T}okens \textbf{E}valuation (CUTE)\footnote{We release our benchmark open-source at: \url{https://github.com/Leukas/CUTE}.}, a benchmark consisting of several tasks designed to be easy for humans to complete, given our ability to process characters %each character in a word 
individually. 
We evaluate %We examine the performance of 
several LLMs ranging from 7B to 132B parameters in size
on CUTE %. With our benchmark, we aim 
to answer the following questions: %\vspace{-3pt}
\begin{enumerate}
    \itemsep-5pt
    \item Do LLMs know which characters make up their tokens?
    \item Do LLMs understand the difference between semantic and orthographic similarity?
    \item Can LLMs manipulate text at the character level?
\end{enumerate} \vspace{-3pt}

\noindent We address these questions with a set of tasks primarily on the character level, and additionally on the word level 
to see the difference in performance on the two granularities, thereby separating the understanding of the task from the understanding of what makes up a token.

\section{Related Work}
% If LLMs do not understand their tokens, they are certainly good at hiding it. LLMs perform well on several tasks which require character-level information, % todo what tasks and cite

\citet{itzhak-levy-2022-models} mostly analyze encoder-only models and test if they understand how to spell words after fine-tuning on 32k examples. They conclude that models learn to spell their tokens ``to some extent.'' They also experiment with GPT-2 \cite{radford2019language} which performed similarly to the other models, and also used %and still required
%HS: Was the performance bad without fine-tuning or didn't they check this? 
% LE: They didnt check this
training examples. \citet{kaushal-mahowald-2022-tokens} probe models with a task asking if a letter is in a word (similar to our \texttt{character contains} task, see \cref{sect:benchmark}). Similar to \citet{itzhak-levy-2022-models}, their probe requires training, %to function,
as they use models of similar size. By contrast, we examine models with 10 to 200 times %10$\times$ to 200$\times$ 
as many parameters and apply few-shot prompting without fine-tuning.
% models roughly 10$\times$ to 200$\times$ larger in parameter count, which also enables us to perform few-shot prompting to test the models as they are.

\citet{huang-etal-2023-inducing} 
experiment with spelling correction, unscrambling words, and finding words in a string of characters.  
Most of their experiments concern a training method they propose, but they also include results of GPT-3 \cite{brown2020language} with few-shot prompting, finding that it performs well on spelling correction, but poorly  on other tasks compared to their trained character-based models.
%They find that with interchange intervention training, they can achieve good performance from subword models on these tasks, though still less performant than character level models. Their approach also requires further training of the models. 
Most of their tasks require semantic knowledge, which we wish to ablate in our benchmark.

Other benchmarks testing orthographic knowledge often focus on morphology, such as the SIGMORPHON inflection tasks % various morphological inflection tasks held by SIGMORPHON 
(e.g. \citet{goldman-etal-2023-sigmorphon}). % todo cite
%These tasks modify words based on a given morphological tag. 
Inflection is fairly regular %rules follow a fairly regular pattern 
%HS: I think your argument here needs clarification. Do you want to say that LLMs might simply memorize all the inflected wordforms which they encountered in their training data without being able to generalize? This could be tested by asking the LLMs to inflect newly created pseudo-words which look like real words. Or are you arguing that LLMs are able to learn inflection rules but fail on other similar tasks which are not represented in the training data? Do you suspect that the pretraining data might include special data which somehow helps the LLMs on inflection tasks, such as a book on English inflection with explicit inflection rules?
% LE: I mean that the inflection patterns are learned, not that they memorize every word in all its forms. But then that does not extend to learning how to manipulate sequences in an arbitrary way. 
for most languages, and could be memorized by a language model. Given that many developers of LLMs do not disclose the sources of their pretraining data, it is possible that LLMs memorize these inflection patterns. Therefore, from this we cannot conclude that LLMs can inflect a new word given a new set of rules, and furthermore we cannot conclude that LLMs can apply an arbitrary manipulation to a sequence.  
Most LLMs are trained with performance on English in mind, and English being less morphological in nature makes inflection a nonideal measure 
for a model's understanding of the composition of tokens.
%to use for gauging whether a model understands the composition of its tokens.

The most similar benchmark to the one we propose is LMentry \cite{efrat-etal-2023-lmentry}. The purpose of LMentry is similar in that the goal is to test models on tasks that are trivial to humans. Some of the tasks included test orthography and are similar to our tasks, for example, they ask the model to write a word containing a letter, or ask to write the first or last letter of a given word. Our benchmark is distinguished from LMentry as most of our tasks explicitly require knowledge of \textit{every} character in a word. Testing whether a model knows the first letter of a word is not sufficient for concluding that it understands orthography, as there may be more pressure to learn about the first letter of a word from pretraining and/or instruction tuning (e.g. via alliterations or acronyms). As such, the task to write a word containing a letter could be more trivially solved by writing a word that starts with said letter. 

There is an extensive body of research 
on character-level models, where each character 
forms a token (\citealp{lee-etal-2017-fully,xue-etal-2022-byt5,tay2022charformer}, \textit{inter alia}).
Several works compared these models to subword models (\citealp{libovicky-etal-2022-dont,edman-etal-2022-subword,edman2024characterlevel}, \textit{inter alia}), but they %though these works have all
evaluated on tasks requiring additional training.
% cite byt5, libovicky, edman, charformer, megabyte
We assume that character-based models would perform well on our benchmark, but we cannot test it since none of these models have been instruction-tuned.
%since there is no character-level instruction-tuned LLM, we are not able to test this.

\begin{figure}
    \centering
    \includegraphics[width=\columnwidth]{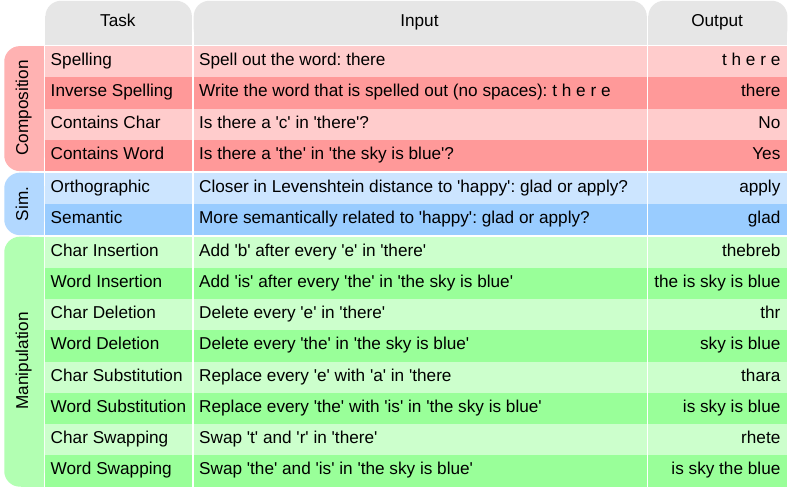}
    \caption{All of the tasks in CUTE.}\label{tab:tasks}
\end{figure}

% \begin{table*}[!htp]\centering
% \scriptsize
% \begin{tabular}{lrrr}\toprule
% \multicolumn{3}{c}{Composition} \\
% Task &Input &Output \\\midrule
% Spelling &Spell out the word: alphabet &a l p h a b e t \\
% Rev spelling &Write the word that is spelled out (no spaces): a l p h a b e t &alphabet \\
% Contains character &Is there a 'c' in 'alphabet'? &No \\
% Contains word &Is there a 'the' in 'the sky is blue'? &Yes \\ \midrule
% \multicolumn{3}{c}{Orthographic and Semantic Similarity} \\ 
% Task &Input &Output \\ \midrule
% Orthographic Similarity &Which is closer in Levenshtein distance to "happy"? glad or apply &apply \\
% Semantic Similarity &Which is more semantically related to "happy"? glad or apply &glad \\ \midrule
% \multicolumn{3}{c}{Manipulation} \\ 
% Task &Input &Output \\ \midrule
% Character insertion &Add 'b' after every 'a' in 'alphabet' &ablphabbet \\
% Word insertion &Add 'is' after every 'the' in 'the sky is blue' &the is sky is blue \\
% Character deletion &Delete every 'a' in 'alphabet' &lphbet \\
% Word deletion &Delete every 'the' in 'the sky is blue' &sky is blue \\
% Character substitution &Replace every 'a' with 'b' in 'alphabet' &blphbbet \\
% Word substitution &Replace every 'the' with 'is' in 'the sky is blue' &is sky is blue \\
% Character swapping &Swap 'a' and 'b' in 'back' &abck \\
% Word swapping &Swap 'the' and 'is' in 'the sky is blue' &is sky the blue \\
% \bottomrule
% \end{tabular}
% \caption{All of the tasks in CUTE, with a simplified example.}\label{tab:tasks}
% \end{table*}

\section{Benchmark} \label{sect:benchmark}

We split our tasks into 3 categories: understanding composition, understanding orthographic similarity, and ability to manipulate sequences. 
Figure~\ref{tab:tasks} shows an example for each task.\footnote{The prompts shown here are not the full prompts. See Appendix \ref{app:prompts} for more details.} 
Data gathering and processing details can be found in Appendix~\ref{app:data_proc}. Our tasks are 
synthetically generated %all generated synthetically 
from existing corpora. There are non-synthetically generated datasets  
which partially test our research questions, but these datasets have external factors (e.g. domain and/or language in a translation dataset) that would likely obscure our findings. 
Some of these datasets also % Additionally, such datasets 
might have been leaked into the LLM's pretraining data and been memorized, resulting in an unrealistically good performance.
%could be susceptible to contamination, and as a result the LLMs may overperform due to memorizing a few specific cases of the tasks we test. 

\subsection{Composition}
We start with a %very simple 
straightforward benchmark: spelling. Similar to \citet{itzhak-levy-2022-models}, we include a task where the input is a word given as a single token\footnote{This is in the ideal case. We cannot guarantee every LLM tested uses only a single token for each input, but we minimize the chance of splitting by using frequent words in our task.}, and the output is the same word with spaces in between, so that each character 
becomes a separate token. %receives its own token. 
This is the most straightforward probe to see whether a model has knowledge of the characters forming the tokens. We also add the inverted task (``inverse spelling'') to check if characters can also be mapped to tokens. 

%HS: I have tried to clarify the arguments regarding this task, but I might have gotten something wrong. So, please check.
Another method for assessing a model's understanding of composition is to ask if a token contains a certain character.
If a model managed the previous tasks, we would expect it to succeed here as well.
%Intuitively, if a model succeeds at the previous tasks, it should succeed in this task as well. 
However, a model might not understand the relationship between spelling and membership of characters to a word,
% However, a model might not correctly understand this task, %may not understand the relationship between the spelling of a word and membership of a set, 
so we test this as well. 
We also test whether the LLMs are able to solve the corresponding word-level task (i.e. is a word in a sentence) %We test this on the character level (i.e. is a letter in a word), as well as on the word level (i.e. is a word in a sentence) 
to separate the model's general understanding of 
the task from its ability to solve the task at the character level.
%membership as a general concept from membership of a character in a word.

\subsection{Similarity}
Since the introduction of word2vec \cite{mikolov2013efficient},
language models have typically been trained to predict words from their context. %the standard practice for training language models has been through predicting words based on context. 
The resulting token embeddings mainly reflect the semantic and syntactic similarity of tokens. % are largely based on similarity in meaning of the tokens, and much less orthographic similarity.
Our next tasks examine whether LLMs also comprehend orthographic similarity. %concern this artifact: we ask whether LLMs, which through extensive contextual pretraining are heavily biased towards semantic similarity, can comprehend orthographic similarity. 
We ask which one of two candidate words is orthographically (or semantically) more similar to a given word.
%how orthographically (or semantically) similar a given word is to two candidate words. 
Our candidate words are chosen to be relatively easy to distinguish for a human without explicit knowledge of how to measure orthographic or semantic similarity. For more details, see Appendix \ref{app:data_proc}.

\subsection{Manipulation}
Our previous tasks focus on the understanding of the model. Now, we turn our focus to acting on that understanding.
% %HS: I didn't understand: internal understanding as-is
% %So far, our tasks largely focus on probing internal understanding as-is, which may be 
% susceptible to memorization. 
% %HS: Maybe explain why they are susceptible to memorization and what is different with the next task.
The next tasks involve 4 types of manipulation of the input at the character or word level: Insertion, Deletion, Substitution, and Swapping. We consider these tasks as elementary %``atomic'' 
tasks for modifying a text sequence. %\footnote{Substitution and swapping can be achieved as a combination of insertions and deletions, though due to their regularity in natural language we opt to include them.}

% For the word-based tasks, we use the TinyStories %todo cite
% dataset, which consists of stories written by an LLM in a style appropriate for a 3-4 year old reader. This has the benefit of using simple sentences with a limited vocabulary, maximizing the chances of words taking up an entire token in the models we test, while also ensuring that the complexity of the sentence is not a confounding factor in a model's performance on the tasks. We filter for sentences of length 3-10 words, in order to make the length similar to the number of characters in a word seen in the character-level tasks. 

\paragraph{Insertion}
First we test how well the model can insert an element X after every instance of some element Y in the sequence. 
%HS: Yaaaaay is an instance of duplication not of insertion of a certain letter. Is that a problem?
% LE: I don't think so, since duplication is a special case of insertion
Similar modifications occur when we replicate letters to emphasize a word %We can see such a modification in natural language with interjections to put additional stress 
(e.g. ``Yay!'' vs. ``Yaaaaay!''), or 
when we add an adjective next to a noun. %, among many other examples. 
% This is a specific case of insertion\footnote{We detail why we did not use insertion in Appendix A.}, requiring the model to insert an additional element next to a target element. We can see such a modification in natural language with interjections to put additional stress (e.g. ``Yay!'' vs. ``Yaaaaay!''), or similarly to add stress to a modifier (e.g. ``I am very very tired'').

\paragraph{Deletion}
Deletion requires the model to recognize an element and remove all instances of it. This can occur in natural language at the character level with 
%HS: Why "synthetic"? Because natural languages always have exceptions? The example is English.
inflection in languages (e.g. turning an English plural noun into singular), %(e.g. making an English plural noun singular), 
or removing adjectives from a sentence. %, as deletion is simply the inverse operation of insertion.

\paragraph{Substitution}
Substitution replaces all instances of an element in a sequence with another element. This can occur with spelling or vocabulary variations across dialects or related languages (e.g. ``defense'' vs. ''defence'', or ``elevator'' vs. ``lift'').

\paragraph{Swapping}
Swapping is a simplified case of reordering acting on two elements.\footnote{Due to the poor performance on swapping, we leave out more complex forms of reordering, but these could be easily added in the future.} Though reordering is not very common in English, it features heavily in languages with free word order, such as Greek, where stressed words can be moved to the front of the sentence. 

% todo when does swapping or reordering occur on the character level?

% \begin{table}[!htp]\centering
% \scriptsize
% \begin{tabular}{lrrrr}\toprule
% &Params &Tokens &Language \\\midrule
% \multirow{3}{*}{Llama} &7B &\multirow{3}{*}{32k} &\multirow{3}{*}{English} \\
% &13B & & \\
% &70B & & \\ \midrule
% \multirow{2}{*}{Mistral / Mixtral} &7B &\multirow{2}{*}{32k} &\multirow{2}{*}{English} \\
% &47B & & \\ \midrule
% Gemma &7B &256k & Hybrid \\ \midrule
% \multirow{2}{*}{Command-R(+)} &35B &\multirow{2}{*}{256k} &\multirow{2}{*}{Multilingual} \\
% &104B & & \\ \midrule
% DBRX &132B &100k &English \\
% \bottomrule
% \end{tabular}

% \end{table}

\begin{table}[!htp]\centering
\scriptsize
% \begin{tabular}{lrrrrrr}\toprule
% &Llama &Mistral &Gemma &DBRX \\\midrule
% Params (B) & 7,13,70 &7,47 &7 &132 \\
% Tokens (k) &32 &32 &256 &100 \\
% % Language &EN &EN &Hybrid &Multil. &EN \\
% \bottomrule
% \toprule
% &Cmd-R(+) &Aya \\\midrule
% Params (B) & 8,35 &35,104 && \\
% Tokens (k) &256 &256 && \\
% % Language &EN &EN &Hybrid &Multil. &EN \\
% \bottomrule

\begin{tabular}{lrrr} \toprule
    Model & Params (B) & Tokens (k) & Lang \\ \midrule
    Llama 2 & 7, 13, 70 & 32 & EN \\
    Gemma & 7 & 256 & EN \\
    Mistral & 7, 47 & 32 & EN \\
    Aya 23 & 8, 35 & 256 & Multil. \\
    Cmd-R(+) & 35, 104 & 256 & Multil. \\
    DBRX & 132 & 100 & EN \\
    Llama 3 & 8, 70 & 100 & EN \\
    \bottomrule
\end{tabular} 

% \end{tabular}
\caption{Models evaluated on our benchmark. We note that Gemma has a multilingual tokenizer, but English-centric training.}\label{tab:models}
\end{table}
\vspace*{-0.3cm}

\section{Experimental Setup}

\paragraph{Models}
We use the models shown in Table \ref{tab:models}.\footnote{We link all of the models in Appendix \ref{app:models}.} We choose freely-available\footnote{We define ``freely available'' as those not requiring payment for use, and with available information on the training process.} LLMs largely based on their popularity, as we are unaware of LLMs that specifically address the problems raised. While there are many differences between the models, we highlight 3 that could possibly affect performance: parameter count, vocabulary size, and multilingual versus English-centric or English-only training.

\paragraph{Prompts}
We use a template inspired by \citet{bsharat2023principled}'s few-shot template to prompt our models with 4 in-context examples. 
Further details and a sample prompt can be found in Appendix \ref{app:prompts}.

\paragraph{Russian Experiments}
We additionally create CUTE-Rus, a Russian version of our benchmark. We discuss this fully in Appendix \ref{app:russian}. The trends largely follow what we see for English.

\begin{figure*}
    \centering
    \includegraphics[width=\textwidth]{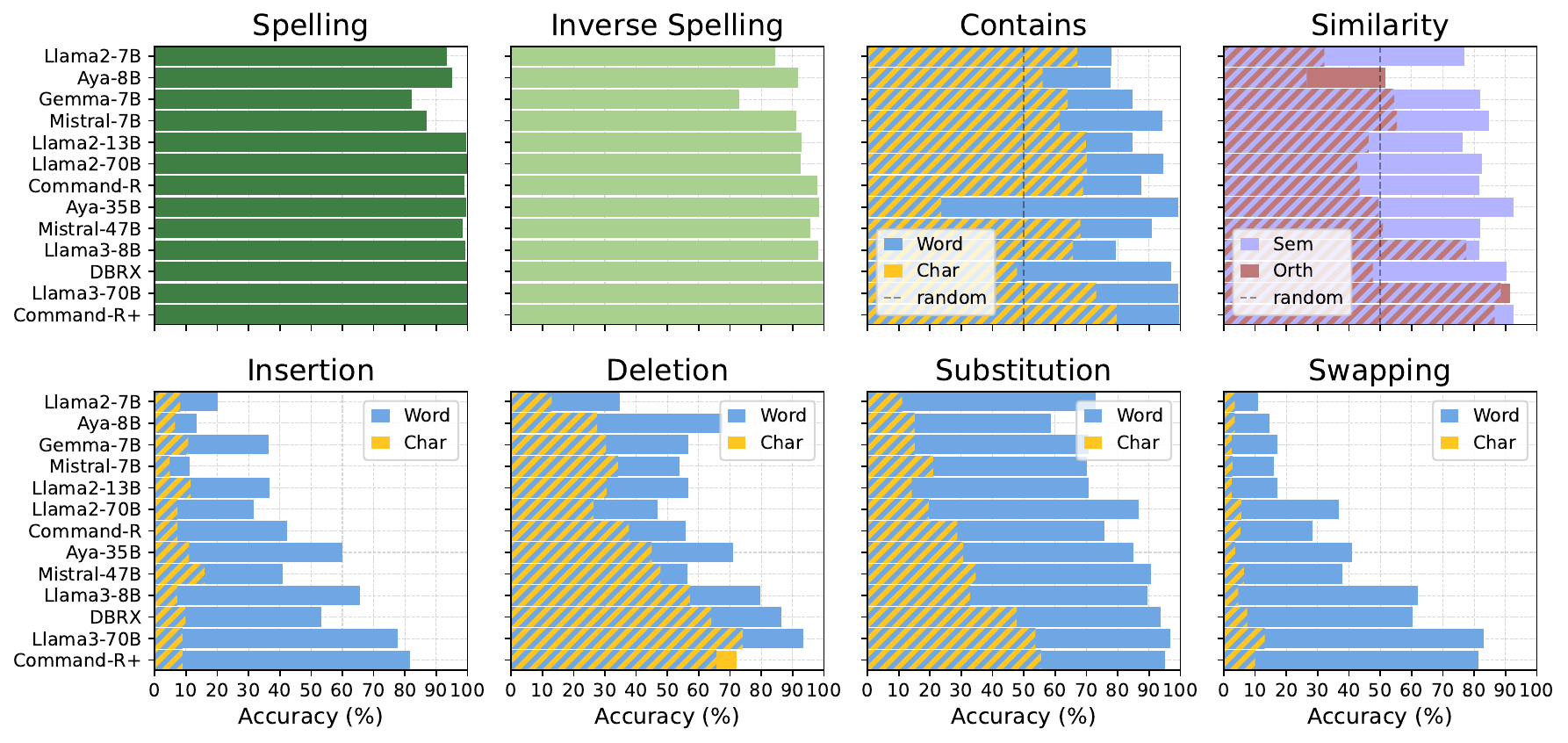}
    \caption{Accuracy on each task of CUTE. Models ordered by average accuracy over all tasks.}
    \label{fig:results}
\end{figure*}

\section{Results}
Figure \ref{fig:results} shows the results for each model. The random baseline is 50\% for the \texttt{contains} and \texttt{similarity} tasks, and 0\% for all other tasks.

\subsection{Composition} 
The models perform very well on the tasks \texttt{spelling} and \texttt{inverse spelling}, though \texttt{inverse spelling} appears slightly more difficult. Although we are not aware of any spelling tasks in instruction tuning or pretraining datasets, we suspect that the similarity of this task with data seen during training allows the models to perform very well, in contrast to the following tasks. 

On the \texttt{contains} tasks, the performance at the word level is quite good, showing that the models understand the task, but the performance breaks down at the character level. This indicates the models do not fully understand the relationship between spelling and membership of a character to a word. 

\subsection{Orthography and Semantic Similarity}
In the \texttt{semantic similarity} task, the models correctly choose the more semantically related word 76-93\% of the time (with the exception of Aya-8B), and the performance generally increases with model size. For \texttt{orthographic similarity}, the performance is below or near random for all models except 
Command-R+ and Llama3. It is not yet clear why they perform so well on this task, but apparently it is not solely a scaling effect since DBRX fails to perform above random chance. It may 
be due to the amount of training data, however DBRX and Command-R+ have not disclosed these amounts. %as compressing the representation of similar languages encourages embeddings which are also orthographically similar.
%HS: Maybe the last point needs additional arguments to support it.

\subsection{Manipulation}
In our manipulation tasks, the models struggle more at the character level than at the word level. The difference is quite profound, with performance gaps of up to 72.8\% on Command-R+ for \texttt{insertion}. Larger models such as Command-R+ perform well on deleting characters, with 72\% accuracy, though it should be noted that we are only testing on the 1000 most frequent words, making the evaluation fairly generous. %even this is a fairly generous evaluation. 

Like the \texttt{contains} task and \texttt{orthographic similarity} task, the manipulation tasks show that LLMs lack a complete understanding of their tokens, although they can literally spell them out. The higher word-level performance indicates that it is not due to a lack of understanding of the task itself. 

\subsection{Vocabulary Size and Multilinguality}
There appear to be no noticeable effects of vocabulary size from the results shown. Looking at the 7/8B models, while Llama 3 performs well with a vocabulary size of 100k, using a larger vocabulary (i.e. Gemma) does not improve performance, and neither does using a smaller vocabulary (i.e. Llama 2 and Mistral). It remains to be seen if noticeable effects arise as the vocabulary size approaches the number of characters. We leave this for future research. 

For multilinguality, the results are similarly mixed. We focus on Aya-35B versus Command-R, as Aya is a fine-tuned version of Command-R, using multilingual instruction tuning data. Overall, Aya makes slight improvements over Command-R on the character-level, but this could easily be due to training on additional English data, rather than the additional non-English data.

% To see the effects of vocabulary size and multilingual versus English tokenization, we compare Llama 2 and 3, along Mistral with Gemma, as they all have 7/8B variants. While Llama 3 performs best and has a larger vocabulary size than Llama 2, Gemma's performance 

% The results are mixed: Mistral and Gemma perform similarly, with Mistral performing slightly better on most tasks, whereas Gemma performed remarkably well on the \texttt{insertion} tasks. We see no obvious connection between the \texttt{insertion} tasks and multilinguality or vocabulary size. 

% Neither vocabulary size nor multilinguality seem to play a huge role. 

\subsection{Scaling}
There are 2 major factors to scale: parameter count and training data. In terms of parameter count, larger models clearly tend to perform better. With respect to amount of training data, only Llama 2 and 3 have disclosed this, and based on the results, it appears that more training data also improves performance. This aligns well with the myriad of works showing the benefits of scaling and raises the question: Is scaling all we need for good performance on character-level tasks? Looking at the manipulation tasks, it seems that \texttt{deletion} and \texttt{substitution} could become manageable in the near future, but for \texttt{insertion} and \texttt{swapping}, the performance gap between word and character level tasks is large. Many real-world text manipulation tasks are a combination of the tested tasks, so we will likely need more than just scaling. 
%In Appendix \ref{app:difficulty}, we find that LLMs struggle more on manipulation tasks 
%We also see that Command-R+'s performance is generally stronger than DBRX, showing that parameter count is not the only factor that can contribute to better performance. %As for the orthographic similarity task, there is no improvement in performance with respect to scaling, so a novel approach to this problem will be necessary. 

% Quite good at spelling, maybe it is in the training data somehow

% They can manipulate text on the word level, but not the character level

\section{Conclusion}
While current LLMs with BPE vocabularies lack direct access to a token's characters, they perform well on some tasks requiring this information, but perform poorly on others. 
The models seem to understand the composition of their tokens in direct probing, but mostly fail to understand the concept of orthographic similarity.
Their performance on text manipulation tasks at the character level lags far behind their performance at the word level. 
LLM developers currently apply no methods which specifically address these issues (to our knowledge), and so we recommend more research to better master orthography.
Character-level models are a promising direction. With instruction tuning, they might provide a solution to many of the shortcomings exposed by our CUTE benchmark.

\section{Limitations}
We prompt instruction-tuned LLMs without any fine-tuning on benchmark data.
While this can be seen as a limitation, we note that it is not feasible to add more training data whenever we discover a new issue with LLMs. 
We expect that the performance of all models would increase after fine-tuning.

We do not evaluate any character-level models since there are no instruction-tuned versions (to our knowledge). Additionally, there are no decoder-only pretrained LLMs available, with the closest model being ByT5-XXL (13B), which has a heavy encoder and smaller decoder. Training character-level models also requires a much higher computational budget, as the sequence lengths are roughly 5 times longer, resulting in 5 times longer training. As such, training a truly comparable model falls outside the scope of this work.

Our benchmark does not control whether LLM tokenizers split words into multiple tokens. We minimize that chance by choosing frequent words, but we can never guarantee that future models will not split words into multiple tokens. We found that the impact of removing split tokens is minimal, with less than 1\% change on average (see Appendix \ref{app:splitting}). 
%which are likely to be tokenized as single tokens, but there is no guarantee that splits do not occur.  

We only test on English and Russian, with our primary focus being on English. While we did not see any major differences in the LLMs' performance between English and Russian when it comes to character-level versus word-level performance, it is possible that there may be differences in other languages. 

% We only test on English, mainly due to the abundance of English-centric LLMs. We expect the results to vary widely with language. 
% Given a vocabulary of frequent words, we could easily create
% \texttt{spelling}, \texttt{inverse spelling}, \texttt{character contains}, and character-level manipulation tasks for other languages. 
% For the similarity tasks, we would need a word embedding model, and for the \texttt{word contains} and word-level manipulation tasks, we would need a small corpus of simple sentences containing mostly frequent words, or we could generate such sentences with an LLM for that language, following TinyStories \cite{eldan2023tinystories}. 

Lastly, we do not control for generations that do not match the pattern of the examples given in the prompt. Therefore, we cannot guarantee that 
all generations considered correct by humans are evaluated as such. 
Hence the performance of some models may be lower than expected. We provide the outputs of all models in our repository for further analysis. 

\section{Acknowledgments}
The work was supported by the European Research Council (ERC) under the European Union's Horizon Europe research and innovation programme (grant agreement No. 101113091) and by the German Research Foundation (DFG; grant FR 2829/7-1). We also thank Lisa Bylinina for helping with the Russian translation of our benchmark.

\bibliography{custom,anthology}

\appendix

\section{All Models} \label{app:models}
Here we link to all of the models. Those associated with published works are as follows:
\citet{touvron2023llama,jiang2023mistral,jiang2024mixtral,gemmateam2024gemma,aryabumi2024aya}.
All models can be found at the following links:
\begin{itemize}
    \itemsep-5pt
    \item \url{https://hf.co/meta-llama/Llama-2-7b-chat-hf}
    \item \url{https://hf.co/meta-llama/Llama-2-13b-chat-hf}
    \item \url{https://hf.co/meta-llama/Llama-2-70b-chat-hf}
    \item \url{https://hf.co/google/gemma-7b-it}
    \item \url{https://hf.co/mistralai/Mistral-7B-Instruct-v0.2}
    \item \url{https://hf.co/mistralai/Mixtral-8x7B-Instruct-v0.1}
    \item \url{https://hf.co/CohereForAI/aya-23-8B}
    \item \url{https://hf.co/CohereForAI/aya-23-35B}
    \item \url{https://hf.co/CohereForAI/c4ai-command-r-v01}
    \item \url{https://hf.co/CohereForAI/c4ai-command-r-plus}
    \item \url{https://hf.co/databricks/dbrx-instruct}
    \item \url{https://hf.co/meta-llama/Meta-Llama-3-8B}
    \item \url{https://hf.co/meta-llama/Meta-Llama-3-70B}
\end{itemize}

\section{Prompting Details} \label{app:prompts}
We show an example of a full prompt in Figure \ref{fig:prompt_ex}. All of our prompts are available with the release of our benchmark. For generation, we use greedy search.
\begin{figure}
    \centering
\begin{tcolorbox}
\texttt{[INST] Spell out the word, putting spaces between each letter, based on the following examples:\\
\\
1. Spell out the word ``alphabet''.\\ Answer: ``a l p h a b e t''\\
2. Spell out the word ``hello''.\\ Answer: ``h e l l o''\\
3. Spell out the word ``zebra''.\\ Answer: ``z e b r a''\\
4. Spell out the word ``tongue''.\\ Answer: ``t o n g u e''\\
\\
\raggedright
Question: Spell out the word ``cow''. [/INST]\\
Answer: ``}
\end{tcolorbox}
    \caption{An example of a full prompt to spell the word ``cow'', with examples, for the task \texttt{spelling}.}
    \label{fig:prompt_ex}
\end{figure}

For evaluation of the generation, we rely on the given start-quote to denote the start of the answer, and we filter out anything after the end quote (e.g. ``I hope this answer helped!''). Some models also were prone to starting generation with a generic response such as ``Sure I can do that for you.''. For these generations, we observe that it would repeat ``Answer: '', so we filter out all generations before this point. Of the remaining generations, some could be considered correct though did not match the desired pattern (e.g. ``H-E-L-L-O'' rather than ``h e l l o'' for the spelling task). These we ultimately consider incorrect so as not to unfairly elevate any model's performance.

Concerning the wording of the orthographic similarity task, it could be argued that models do not understand the concept of Levenshtein distance, and thus it is not comparable to the semantic similarity task. We also tested using ``closer in edit distance'' and ``closer in spelling'' in the prompt, and the results were very similar, so we opted for Levenshtein distance as it is more well-defined. Similarly, we used ``closer in meaning'' rather than ``more semantically related'' and achieved similar results, though it is debatable whether an antonym should be considered close in meaning, so we opted for the latter. 

\begin{figure*}[!ht]
    \centering
    \includegraphics[width=\textwidth]{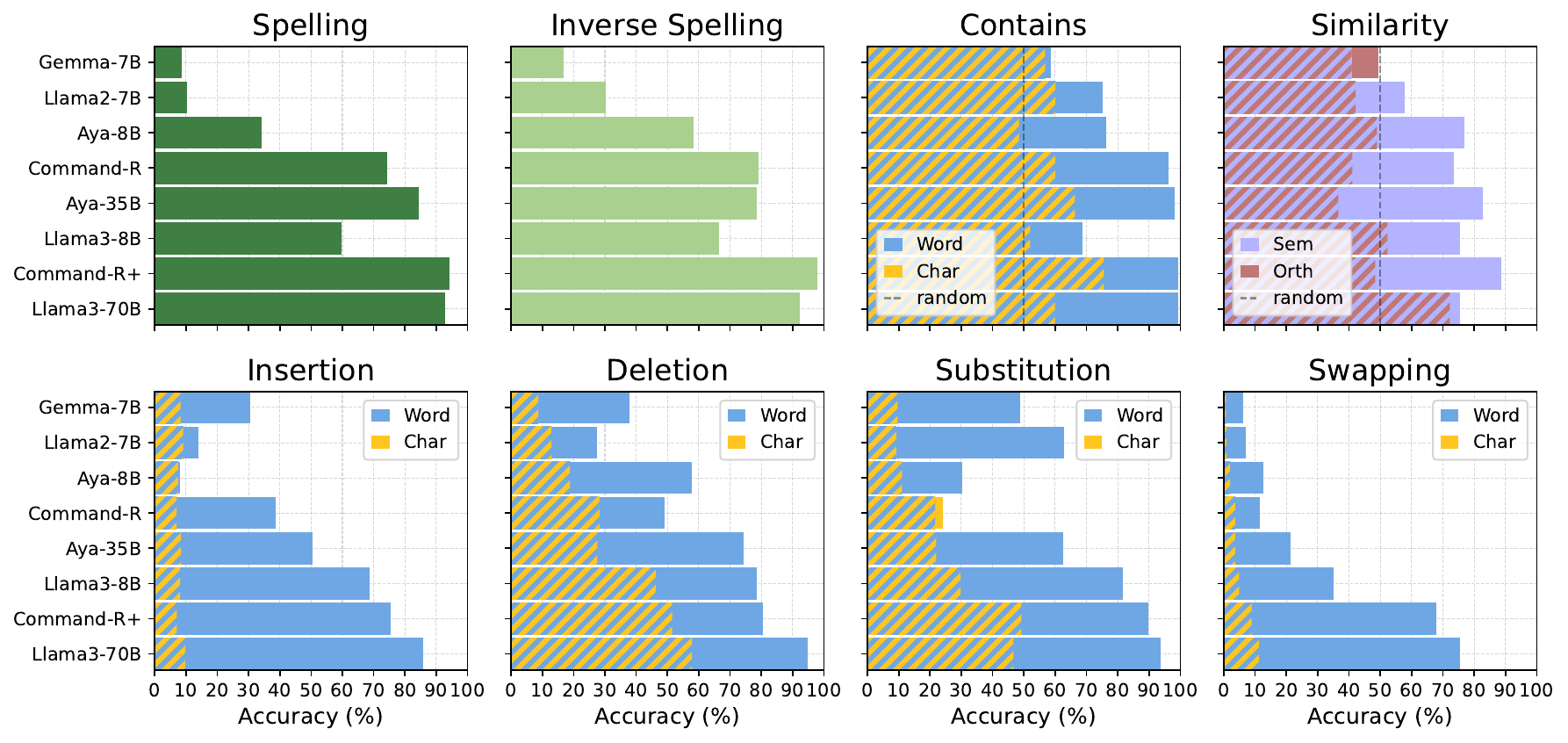}
    \caption{Accuracy on each task of CUTE-Rus. Models ordered by average accuracy over all tasks.}
    \label{fig:results_ru}
\end{figure*}

\section{Data Processing} \label{app:data_proc}
Here we detail our exact method for gathering and processing the data into our tasks. The scripts for processing the data and the resulting data can be found at our Github repository.\footnote{\url{https://github.com/Leukas/CUTE}} 

\paragraph{Data Sources}
For almost all tasks, we require a set of frequent English words that were most likely to be tokenized into a single token. For this, we use a dataset derived from the Google Web Trillion Word Corpus.\footnote{\url{https://www.kaggle.com/datasets/rtatman/english-word-frequency/data}}

For the word-based tasks, we use the TinyStories \cite{eldan2023tinystories} dataset, which consists of stories written by an LLM in a style appropriate for a 3-4 year old reader. This has the benefit of using simple sentences with a limited vocabulary, maximizing the chances of words being tokenized into a single token in the models we test, while also ensuring that the complexity of the sentence is not a confounding factor in a model's performance on the tasks. 

\paragraph{Filtering}
For character-based tasks, we select the 1000 most frequent words that are at least 3 characters long.
For word-based tasks, we similarly filter for 1000 sentences of length 3-10 words, in order to make the length similar to the number of characters in a word seen in the character-level tasks. 

For insertion, deletion, and substitution, we 
apply the modification to those 1000 words, resulting in our dataset. 

For swapping, we need to sure that the word or sentence has 2 items that are unique, so as to avoid an ambiguous prompt (e.g. swap the `e' and `g' in `engineering'). As such, we select the 1000 most frequent words or the first 1000 sentences that satisfy this criteria, as well as satisfying our length constraints.

\paragraph{Similarity Data}
For our similarity data, we require our candidate pairs to be sufficiently easy for a human to distinguish which is closer orthographically and which is closer semantically. 

To accomplish this, our candidate words must satisfy two thresholds, one based on normalized Levenshtein distance (for othographic similarity), and one based on cosine similarity to other fastText \cite{bojanowski-etal-2017-enriching}
embeddings (for semantic similarity). That is to say, the word must be sufficiently similar in one metric (0.7+ and 0.5+ for Levenshtein and cosine, respectively) and sufficiently dissimilar in the other ($0.3-$ and $0.2-$, respectively). These thresholds are decided empirically. We note that this process occasionally ends up with semantic pairs that are antonyms (e.g. ``good'' and ``bad''), and thus we refrain from stating that the pairs are similar in meaning. 

% What we consider to be ``sufficiently'' similar or dissimilar is decided empirically, though we provide positive and negative examples below:

% \begin{table}[!htp]
%     \centering
%     \begin{tabular}{lrrr}
%       Similarity & Word & Sufficient & Insufficient \\
%        Orthographic & scab & slab & scamp \\
%        Semantic & scab & slab & scamp        
%     \end{tabular}
%     \caption{Target words and their sufficient or insufficient candidates.}
%     \label{tab:my_label}
% \end{table}

\begin{figure*}[!ht]
    \centering
    \includegraphics[width=\textwidth]{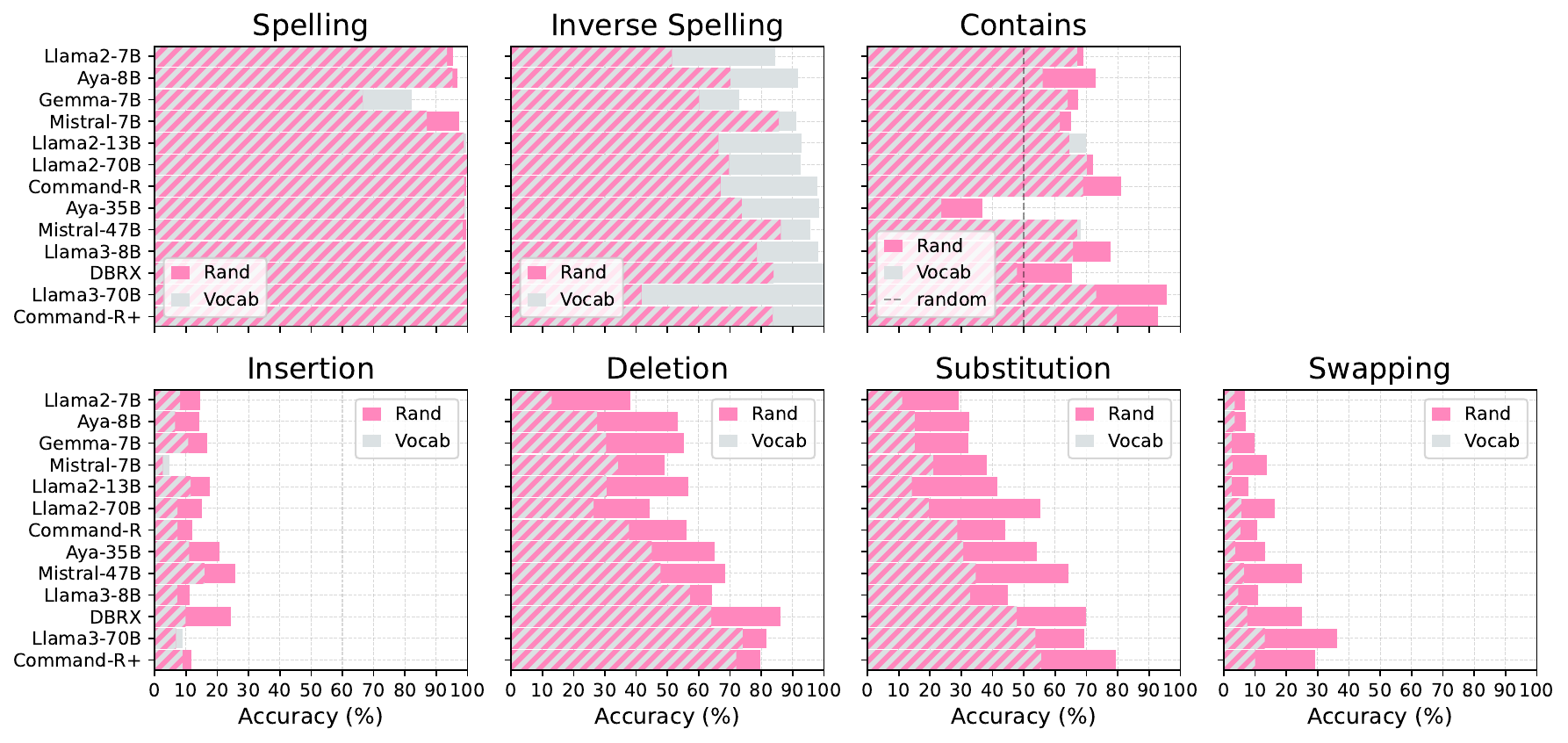}
    \caption{Evaluation of the performance on random strings versus strings in the vocabulary. ``Vocab'' is equivalent to the character-level tasks from Figure \ref{fig:results}.}
    \label{fig:rand_string}
\end{figure*}

\section{Russian Experiments} \label{app:russian}
Here, we detail the creation and evaluation on our Russian version of CUTE, CUTE-Rus. 

\paragraph{Data Processing}
Our data processing followed our processing in Appendix \ref{app:data_proc}. We collected a frequency list of Russian words from Wiktionary\footnote{\url{https://w.wiki/AupG}}%\footnote{\url{https://ru.wiktionary.org/wiki/\%D0\%9F\%D1\%80\%D0\%B8\%D0\%BB\%D0\%BE\%D0\%B6\%D0\%B5\%D0\%BD\%D0\%B8\%D0\%B5:\%D0\%A1\%D0\%BF\%D0\%B8\%D1\%81\%D0\%BE\%D0\%BA_\%D1\%87\%D0\%B0\%D1\%81\%D1\%82\%D0\%BE\%D1\%82\%D0\%BD\%D0\%BE\%D1\%81\%D1\%82\%D0\%B8_\%D0\%BF\%D0\%BE_\%D0\%9D\%D0\%9A\%D0\%A0\%D0\%AF:_\%D0\%A3\%D1\%81\%D1\%82\%D0\%BD\%D0\%B0\%D1\%8F_\%D1\%80\%D0\%B5\%D1\%87\%D1\%8C_1\%E2\%80\%941000}}
% \footnote{\foreignlanguage{russian}{\url{https://ru.wiktionary.org/wiki/Приложение:Список_частотности_по_НКРЯ:_Устная_речь_1—1000}}}
, embeddings from FastText, and we translated the English TinyStories dataset using Google Translate. % TODO cite google translate
We adjusted the thresholds for gathering our orthographic and semantic similarity pairs to (0.55+, 0.55+) and (0.3-, 0.1-) for Levenshtein and cosine, respectively.
The thresholds were more difficult to adjust without excluding too many pairs, and as a result, some of the triplets could be argued as unclear or incorrect. For example, for the triplet 
(\foreignlanguage{russian}{общий}, \foreignlanguage{russian}{община}, \foreignlanguage{russian}{совместный}) meaning (\textit{common}, \textit{community}, \textit{joint}), it is less clear which of the last two are more semantically related to the first (more so in Russian), though \textit{joint} is the intended semantically-related word.

The prompts were also machine translated with Google Translate and post-edited by a native Russian and fluent English speaker, but after testing with both English and Russian prompts, we found the models generally performed better with English prompts (with Russian examples in the prompt), so we only include those in our results.

\paragraph{Results}
We test a subset of the models used in the English version, focusing mainly on the multilingual LLMs (Aya, Command-R(+), and Gemma to the extent of the tokenizer), as well as Llama 2 and 3 for reference. 

Figure \ref{fig:results_ru} shows a similar trend to the English results. Generally speaking, the character-level performance lags behind the word-level. An interesting difference is that the models struggle much more with spelling. Even though the prompt is in English and shows examples of Russian words being spelled out, this is not enough for most models to understand the concept of spelling. The additional multilingual instruction tuning done for training Aya appears to be necessary for better performance.  

% \section{Manipulation Difficulty} \label{app:difficulty}
% To gauge the difficulty of the manipulation tasks, we stratified our results based on the number of characters that needed to be manipulated for the \texttt{deletion} and \texttt{substitution} tasks. 

\section{Random String Evaluation} \label{app:rand}
% TODO decide whether this is useful to see or not
While we cannot directly assess the performance of character-level LLMs without training an equivalent model from scratch, we can evaluate the performance of the existing models when the number of tokens per word approaches the number of characters per word. We can do this by conducting our tasks using random strings of consonants rather than complete words. 

Since practically every word in English requires a vowel, random sequences of consonants are typically quite rare, and thus BPE will not dedicate a singular token for sequences like ``fxqg''. As such, we generate random strings for use in our tasks (excepting the similarity tasks). The resulting strings use on average 1.6 characters per token, compared to 5.4 characters per token in the original word list.  

In Figure \ref{fig:rand_string}, we can see the performance of the LLMs on the character level tasks using regular words (as shown before in Figure \ref{fig:results}), as well as random strings. We can see that, apart from \texttt{inverse spelling}, the models perform the same or better on random strings than on actual words. These tasks appear much easier for LLMs to handle when their tokenization is close to character-level, suggesting that a truly character-level LLM would perform the best. 

As for \texttt{inverse spelling}, the decrease may be a result of the models' bias towards generating words. Upon inspection of the outputs, we observe that occasionally the model would hallucinate a word, changing a string such as ``c q n r w'' to ``conquer'', essentially filling in what it considers to be the missing vowels. This phenomenon is particularly common in Llama3-70B, whose performance decreased the most.

\section{Token Splitting Impact} \label{app:splitting}
% TODO write about impact of split tokens
We mention that while we use frequent words for evaluation to maximize the chance they are given a single token, we cannot guarantee that some words will not be split into multiple tokens.
Here, we evaluate the impact of this splitting on our results to elucidate whether this issue could affect the trends we see in the paper.

% TODO add table with number of tokens split, acc change +- on all manip tasks

\begin{figure}
    \centering
    \includegraphics[width=\columnwidth]{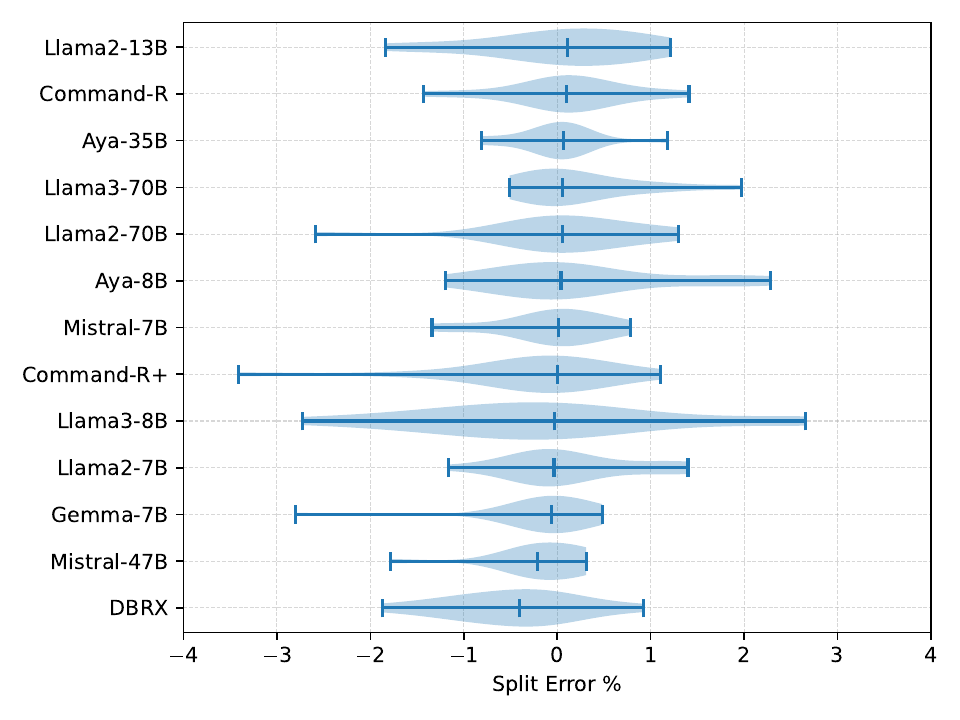}
    \caption{Distribution of change in accuracy for all tasks per model, sorted by median change.}
    \label{fig:split}
\end{figure}

Figure \ref{fig:split} shows the percent change in accuracy were those examples to be removed from each task. All of the tasks are grouped into a violin plot. Here we can see that the maximum accuracy difference is around $-3.5\%$, and the median errors for each model are no greater than $\pm0.5\%$. This is far from substantially affecting the disparity we see between the character-level and word-level tasks. This also reveals that even if an LLM's tokenizer splits a word into two or more tokens, the LLM will still have difficulty performing the tasks in the CUTE benchmark. 

% \section{Difficulties with Training a Character-level LLM}
% As we did not assess a character-level model and also speculate that such a model could perform well on CUTE, we discuss here why training a character-level LLM comes with additional challenges

\end{document}